\documentclass[runningheads]{llncs}

\usepackage{graphicx}
\usepackage{dsfont}
\usepackage{amsmath}
\usepackage{float}
\usepackage{mathtools}
\usepackage[T1]{fontenc}
\usepackage{caption}
\usepackage{subcaption}
\usepackage{textalpha}
\usepackage{multirow}
\usepackage{booktabs} % For formal tables
\usepackage{hhline}
\usepackage{longtable}
\usepackage{multicol}
\usepackage{tabularx}
\usepackage{amssymb}% http://ctan.org/pkg/amssymb
\usepackage{adjustbox}
\usepackage{lipsum}
\usepackage{tikz}
\usepackage{enumitem}
\usepackage{makecell}
\usepackage{xcolor,pifont}
\newcommand{\xmark}{\ding{55}}%
\newcommand*\colourcheque[1]{%
  \expandafter\newcommand\csname #1cheque\endcsname{\textcolor{#1}{\ding{52}}}%
}

\usepackage[colorlinks=true, citecolor=citecolor, linkcolor=linkcolor]{hyperref}
\definecolor{citecolor}{HTML}{0071BC}
\definecolor{linkcolor}{HTML}{ED1C24}
\usepackage{verbatim}
\usepackage[capitalize]{cleveref}

\begin{document}

\title{Enhanced Bank Check Security: Introducing a Novel Dataset and Transformer-Based Approach for Detection and Verification}
\titlerunning{Bank Check Detection and Verification}
\author{Muhammad Saif Ullah Khan*\inst{1,2,3}\orcidID{00090-0002-7375-807X} \and
Tahira Shehzadi*\inst{1,2,3}\orcidID{0000-0002-7052-979X} \and
Rabeya Noor\inst{1,2,3}\orcidID{0009-0005-7867-2143}\and
Didier Stricker\inst{1,2,3} \and
Muhammad Zeshan Afzal\inst{1,2,3}\orcidID{0000-0002-0536-6867}}
\authorrunning{Khan et al.}
% First names are abbreviated in the running head.
% If there are more than two authors, 'et al.' is used.
%
\institute{Department of Computer Science, Technical University of Kaiserslautern, Germany \and
Mindgarage, Technical University of Kaiserslautern, Germany \and
 German Research Institute for Artificial Intelligence (DFKI), 67663 Kaiserslautern, Germany\\
\email{\{muhammad\_saif\_ullah.khan@dfki.de\}}}
\def\thefootnote{*}\footnotetext{These authors contributed equally to this work.}

\maketitle              % typeset the header of the contribution

\begin{abstract}
Automated signature verification on bank checks is critical for fraud prevention and ensuring transaction authenticity. This task is challenging due to the coexistence of signatures with other textual and graphical elements on real-world documents. Verification systems must first detect the signature and then validate its authenticity, a dual challenge often overlooked by current datasets and methodologies focusing only on verification. To address this gap, we introduce a novel dataset specifically designed for signature verification on bank checks. This dataset includes a variety of signature styles embedded within typical check elements, providing a realistic testing ground for advanced detection methods. Moreover, we propose a novel approach for writer-independent signature verification using an object detection network. Our detection-based verification method treats genuine and forged signatures as distinct classes within an object detection framework, effectively handling both detection and verification. We employ a DINO-based network augmented with a dilation module to detect and verify signatures on check images simultaneously. Our approach achieves an AP of 99.2 for genuine and 99.4 for forged signatures, a significant improvement over the DINO baseline, which scored 93.1 and 89.3 for genuine and forged signatures, respectively. This improvement highlights our dilation module's effectiveness in reducing both false positives and negatives. Our results demonstrate substantial advancements in detection-based signature verification technology, offering enhanced security and efficiency in financial document processing.
\end{abstract}

\section{Introduction}
\label{sec:Introduction} 

Signatures remain one of the most widely used behavioral biometrics for authentication systems despite the growing popularity of physiological biometrics such as fingerprints, face scans, and eye scans~\cite{khan2018signature,dargan2020survey}. This is largely because behavioral biometrics are comparatively less invasive~\cite{liang2020behavioral}. However, unlike physiological biometrics which are innate, signatures can be learned and replicated with practice~\cite{sarkar2020review}. Thus, ensuring the security of behavioral authentication systems, such as signature verification, is crucial~\cite{Kao_app101,Vorugunti_FuseNetOS}.

Signatures are important across various fields as a personal mark of consent and authentication. They provide binding validity to documents such as contracts and wills in legal contexts. This importance extends to the financial sector, where signatures are pivotal for authorizing transactions like checks, fund transfers, and account changes, safeguarding against unauthorized access and fraud~\cite{dargan2020survey}.

In banking, signatures on checks are essential for validating transactions and ensuring the legitimacy of financial exchanges. This leads to the complex task of signature verification on bank checks, which involves accurately detecting signatures amidst elements such as text, decorative lines, and logos. These elements often cluster around or overlap with the signature space, complicating the task of signature recognition and requiring advanced methods to separate and identify the signature from its surroundings effectively.

Traditional research in signature verification~\cite{dargan2020survey,pal_sig34} has often focused on simple datasets with signatures on plain backgrounds, aiding the development of algorithms for signature verification~\cite{OKAWA_2018480}. However, in real-world scenarios, particularly with bank checks, signatures frequently appear amidst complex patterns and elements, presenting substantial challenges for accurate verification~\cite{FIERREZ20}. This makes previous approaches~\cite{dargan2020survey,pal_sig34,patel_2017,Narwade_2018OfflineSV} less effective when applied to actual bank checks.

To address this gap, there is a strong need for a dataset that mirrors real-world scenarios where signatures are embedded within a mix of patterns, text, and images, similar to bank checks~\cite{OKAWA_2018480,khan2018signature}. Such a dataset would improve the applicability of signature verification technology in real-world financial and legal contexts.

In this paper, we introduce a new dataset specifically designed for signature verification on bank checks. This dataset accurately reflects the complex environments of bank checks, featuring a variety of signature styles set against common check elements, thus providing a more challenging and realistic testing ground for advanced verification methods. This dataset aims to foster the development of robust and advanced signature verification approaches, enhancing their applicability and reliability in real-world banking operations.

Furthermore, we propose an advanced approach for signature verification using a detection network, tailored for complex overlapping scenarios encountered in banking and legal documents. Our method employs a DINO-based network~\cite{dino23}, augmented with a dilation module to enhance the visibility of thin strokes and improve feature extraction, making the signature's defining characteristics more distinguishable for accurate verification. This approach effectively localizes and verifies signatures on various documents, handling multi-scale features and focusing on relevant areas using deformable attention, thereby addressing the complexities of varying signature styles and backgrounds.

Our results demonstrate significant advancements in the capability of signature verification technologies, offering enhanced security and efficiency in financial document processing. The key contributions of this paper are summarized as follows:
\begin{itemize}
\item[$\bullet$] We introduce the Synthetic Signature Bankcheck Images (SSBI) dataset, featuring real and forged signatures embedded in complex scenarios on bank checks, such as stamps, logos, and background designs, along with other bank check elements. The complete source code and the dataset are available at \url{https://github.com/saifkhichi96/ssbi-dataset/}.

\item[$\bullet$] We present an end-to-end trainable framework based on DINO~\cite{dino23}, incorporating both training and guiding networks for fraud detection on bank checks. This framework effectively handles signature verification and bank check object detection.

\item[$\bullet$] Our approach achieves a significant performance improvement, with a 10\% increase over the baseline DINO network on our dataset. This improvement underscores the efficacy of our method in enhancing signature detection and verification.
\end{itemize}

The remainder of this paper is organized as follows: Section~\ref{sec:Related_Work} provides a thorough review and analysis of the existing datasets available in the field of signature forgery detection. Section~\ref{sec:motivation} explains the motivation behind our research. In Section~\ref{sec:dataset}, we introduce our new dataset, detailing its composition and the processes involved in its creation. Section~\ref{sec:method} elaborates on our approach, and Section~\ref{sec:Experiments} evaluates the generated data for bank check object detection and signature verification, including ablation studies for our proposed approach. Finally, Section~\ref{sec:conclusion} summarizes our key findings and outlines future research directions.

\section{Related Work}
\label{sec:Related_Work}

\subsection{Signature Verification}

Signature verification is important for the banking sector, where handwritten signatures are commonly used~\cite{dargan2020survey}. Online signature verification has access to real-time characteristics of the signature as it is performed on a digital device. In contrast, offline signature verification uses scanned images of a signature, making it a more challenging task~\cite{khan2018signature}. Writer-independent signature verification aims to authenticate a signature regardless of its author. Algorithms in this process search for characteristics commonly seen in forgeries, like shaky strokes and overwriting. On the other hand, writer-dependent signature verification is used when the owner's identity is known, and the objective is to determine if the signer is the owner or not.

\subsubsection{Classification-based Approaches}
In recent years, offline signature verification has seen substantial advancements through many research efforts, each focusing on improving the accuracy and reliability of signature recognition systems. Pal et al.~\cite{pal_sig34} delved into texture features and reported an Average Error Rate (AER) of 32.72\% when utilizing Local Binary Patterns (LBP) and Uniform Local Binary Patterns (ULBP). This study shed light on the significance of texture-based features in signature verification. Patil et al.~\cite{patel_2017} introduced a writer-independent approach by employing a Histogram of Oriented Gradient (HOG) features and a K-Nearest Neighbor (K-NN) classifier, further diversifying the landscape of signature recognition techniques. The research landscape expanded as scholars explored advanced methods for offline signature verification. Fierrez et al.~\cite{FIERREZ20}  harnessed the power of Hidden Markov Models (HMM) and dynamic time functions, achieving remarkable error rates in signature verification. Narwade et al.~\cite{Narwade_2018OfflineSV} introduced shape correspondence and employed a Support Vector Machine (SVM) classifier, showcasing an impressive accuracy of 89.58\% on synthetic signature data. These methodologies collectively highlight the diverse and evolving approaches within offline signature recognition. Moreover, researchers have demonstrated their commitment to robust evaluation by testing their methods across various datasets. Okawa~\cite{OKAWA_2018480} achieved an impressive Equal Error Rate (EER) of 5.47\% on the MCYT-75 dataset~\cite{Fierrez_MCYT-75}, surpassing state-of-the-art systems. Sharif et al.~\cite{SHARIF_202050} adopted genetic algorithm feature selection and SVM classifiers on datasets like MCYT~\cite{Fierrez_MCYT-75} and GPDS~\cite{GPDS_data56}, consistently outperforming existing approaches. These studies underscore the importance of comprehensive evaluation of diverse datasets to validate the effectiveness of signature verification systems. In summary, offline signature verification has undergone substantial progress, with researchers continuously exploring diverse techniques and evaluating their performance on various datasets. These efforts collectively contribute to developing more accurate and dependable signature recognition systems, addressing the growing need for secure and efficient document authentication processes.

\subsubsection{Deep Learning-based Approaches}
Recent advancements in deep learning have broadened their applications, including healthcare~\cite{shehzadi_IEEE_I9,Protein10}, traffic analysis~\cite{wajahatCC8}, to document analysis~\cite{semi-detr_table1,sunsupdla12,shehzadi2024hybrid6,shehzadi2024endtoend7,ehsan_semi8,continuaLR45,Real_DICls4,cas10}. The advancements in signature verification technology have been marked by various innovative deep learning-based approaches~\cite{sparse_semi_detr2,bridging_per3,semimask4}, each contributing significantly to the field. Shariat et al.~\cite{Shariatmadari_2019} introduces a writer-dependent method for signature verification, utilizing a hierarchical one-class Convolutional Neural Network (CNN). This approach is unique in learning authentic signatures without needing forgeries as a reference. The effectiveness of this method is demonstrated on Persian databases (PHBC and UTSig) and Latin databases (MCYT-75 and CEDAR~\cite{CEDAR}), where it outperformed existing state-of-the-art results. Wei et al.~\cite{wei_2019_1} introduce the Inverse Discriminative Networks (IDN) model, designed for writer-independent handwritten signature verification. This model is notable for incorporating four network streams, each analyzing pairs of signature samples. The IDN model's performance is impressive, showing high verification accuracy on a comprehensive Chinese signature dataset and international datasets such as CEDAR~\cite{CEDAR}, BHSig-B, and BHSig-H~\cite{BHSig260}. In 2020, Jain et al.~\cite{Jain_2020HandwrittenSV} developed a shallow Convolutional Neural Network (CNN) approach for verifying handwritten signatures. This method stands out for its language independence, which applies to signatures in any language. This is validated through experiments on various public signature datasets, including the newly created CVBLSig-V1 and CVBLSig-V2~\cite{CVBLSig}, demonstrating its wide applicability and effectiveness in signature verification. Further contributing to this field, Pal et al.~\cite{pal_2016skri} focus on language-independent signature verification and achieve remarkable recognition rates on several datasets, including BHSig260 Hindi and Bengali~\cite{BHSig260} and the MCYT-100 dataset. This method not only surpassed existing methods in recognition accuracy but also demonstrated impressive results in Equal Error Rate (EER) metrics, further solidifying its effectiveness.

Kao at al.~\cite{Kao_app101} brings a new perspective with their deep CNN approach for offline signature verification and forgery detection. This method is particularly effective in scenarios where only a single known signature specimen is available, achieving high accuracy on the ICDAR2011 SigComp dataset. Poddar et al.~\cite{PODDAR_2020610} proposes a deep learning-based approach for offline signature recognition and forgery detection. Combining CNN with the Crest-Trough method for signature recognition and employing the SURF~\cite{BAY2008346} and Harris algorithms~\cite{harris1988combined} for forgery detection, their method showed significant improvement, indicating the effectiveness of this combined approach. Vorugunti et al.~\cite{Vorugunti_FuseNetOS} uses a hybrid approach involving Convolutional Autoencoder (CAE) features and handcrafted features fed into a Depth-wise Separable Convolutional Neural Network (DWSCNN) for lightweight Online Signature Verification (OSV). This method demonstrates promising results on datasets like MCYT-100 and SUSIG. In 2021, Ghosh~\cite{Ghosh_2020ARN} proposes a Recurrent Neural Network (RNN)-based model with LSTM and BLSTM for offline signature recognition and verification. This model is notable for achieving low Equal Error Rates (EERs) on various datasets. Expanding into the realm of 3D signature recognition, Ghosh et al.~\cite{GHOSH_b202113} introduces a Spatio-Temporal Siamese Neural Network (ST-SNN), which performs well on a 3D signature benchmark dataset. Junior et al.~\cite{FCNRLAF7} introduce a novel approach combining Fully Convolutional Networks with Refinement Layers, specifically designed to accurately segment offline handwritten signatures, contributing to improved document processing systems. Lastly, Liu et al.~\cite{Liu2021_OfflineSV} presents a region-based deep metric learning network for offline signature verification. This method is applied to writer-dependent and writer-independent scenarios, achieving competitive Equal Error Rates (EERs) on challenging datasets like CEDAR~\cite{CEDAR} and GPDS. These developments collectively highlight the dynamic and evolving nature of signature verification technology. Each method brings forward new perspectives and solutions, contributing to more secure and reliable systems for signature authentication in various applications.

\subsection{Bank Check Signature Datasets}
Various signature databases are available, such as GPDS~\cite{GPDS_data56}, MCYT~\cite{Fierrez_MCYT-75}, CEDAR~\cite{CEDAR}, BHSig260~\cite{BHSig260}, UTSig~\cite{UTSIg45}, ICDAR2011~\cite{icdar11_sig} and SigComp~\cite{liwicki2011signature} datasets. However, these datasets only contain cropped signatures on white backgrounds, which do not represent signatures on complex documents in real-world scenarios. Verifying signatures on complex documents is challenging as it requires detecting and extracting signatures. Unfortunately, existing signature verification methods tested on these databases do not account for these factors. A small dataset of bank checks, called BCSD, contains signature segmentation annotations, as reported in~\cite{khan2021novel}. However, this dataset lacks any forgeries and has only 156 samples, making it unsuitable for signature verification using deep learning techniques.

\section{Motivation}
\label{sec:motivation}

\noindent This section outlines the primary motivations behind creating our dataset and its anticipated impact on signature verification for bank checks.

\noindent\textbf{Comprehensive Collection of Signature Data from Bank Checks:}  Our dataset is meticulously designed to capture a wide variety of signature styles, including variations in handwriting and ink properties, against diverse backgrounds such as different paper textures, colors, and printed patterns found on checks. This comprehensive approach ensures the dataset is robust and representative of real-world banking scenarios. It prepares algorithms to tackle challenges in signature detection, such as deciphering signatures overlaid on complex backgrounds mixed with printed text and other markings.
%\noindent\textbf{Targeted for Advanced Detection Approaches:} The primary aim of this dataset is to enable the development and enhancement of cutting-edge signature detection algorithms. These algorithms are crucial for automating the signature verification process traditionally done by humans. Training models with this diverse dataset will help accurately distinguish between genuine and forged signatures, detect obscured or distorted signatures, and efficiently and reliably process a high volume of checks. The complexity and depth of the dataset push the boundaries of what automated systems can achieve in signature verification, enhancing their reliability in high-stakes financial environments.

\noindent\textbf{Enhancing Financial Security and Operational Efficiency:} Our dataset significantly contributes to financial security by enabling high-accuracy automatic signature verification, a crucial need for processing financial documents like checks. Reducing reliance on manual verification strengthens security measures against fraud and forgery and boosts operational efficiency. Financial institutions can process more checks faster, improving customer satisfaction with quicker processing times and enhanced security measures. This advancement marks a significant step in the digital transformation of financial services, leading to more secure and efficient banking operations.

%\noindent\textbf{Addressing Challenges in Signature Verification:}
%Focusing on real-world bank check signatures, our dataset addresses the sensitive nature of banking documents, which often contain private information and are generally unavailable for public research. By centering on bank check signatures—which typically do not enclose sensitive personal information beyond the signature itself—we create a dataset that can simulate real-life fraud scenarios like forged signatures for unauthorized withdrawals or altered signatures for financial fraud. This focus ensures that our signature verification methods are tested against scenarios they are likely to encounter in practical banking environments, thus enhancing their effectiveness and applicability.

\begin{figure}[ht]
    \centering
    \includegraphics[width=\linewidth]{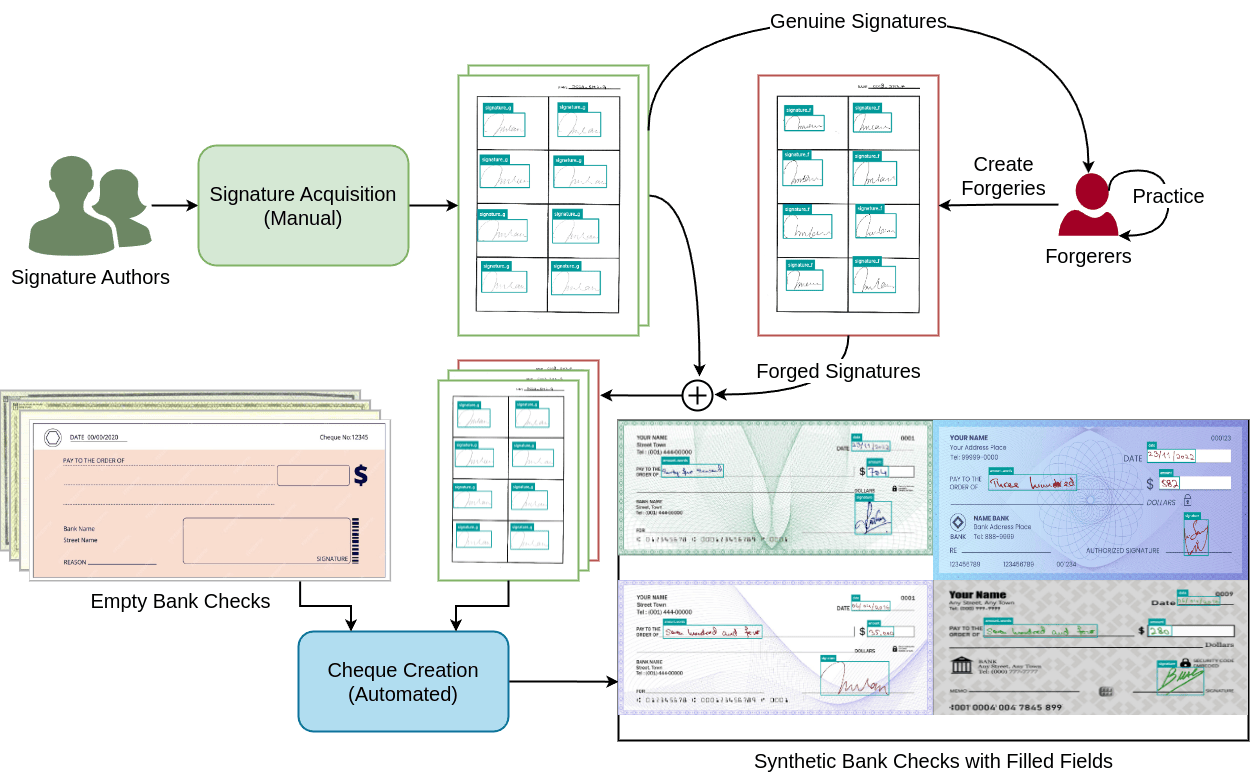}
    \caption{\textbf{Dataset Creation Pipeline.} We use a semi-automated process for dataset generation comprising of a manual signature acquisition step--for obtaining forged and genuine signatures--and an automated step for inserting these signatures and filling other fields on bank checks.}
    \label{fig:data_creation_pipeline}
\end{figure}

\section{The SSBI Dataset}
\label{sec:dataset}

We present the Synthetic Signature Bankcheck Images (SSBI) dataset, featuring diverse signature styles and background complexities to simulate real-world scenarios effectively. This section describes the data collection, preprocessing, and annotation processes that ensure the dataset's relevance and applicability to fraud detection in banking. The complete data creation pipeline is illustrated in Fig.~\ref{fig:data_creation_pipeline}.

\subsection{Signature Data Acquisition}

As the first step of our dataset creation pipeline (Fig.~\ref{fig:data_creation_pipeline}), we collected authentic signatures from real individuals. We designed a standard signature collection sheet with a 4x2 grid to achieve this and requested 19 people to sign their names multiple times. The participants were of different genders, aged between 20 and 40, and had a high-education background. Each participant signed their name on approximately two sheets containing 16 signatures. Half of the signatures were signed with a black ballpoint pen, and the other half with a lead pencil.

\begin{figure}
    \centering
    \begin{subfigure}[b]{0.325\textwidth}
         \centering
         \includegraphics[width=\textwidth]{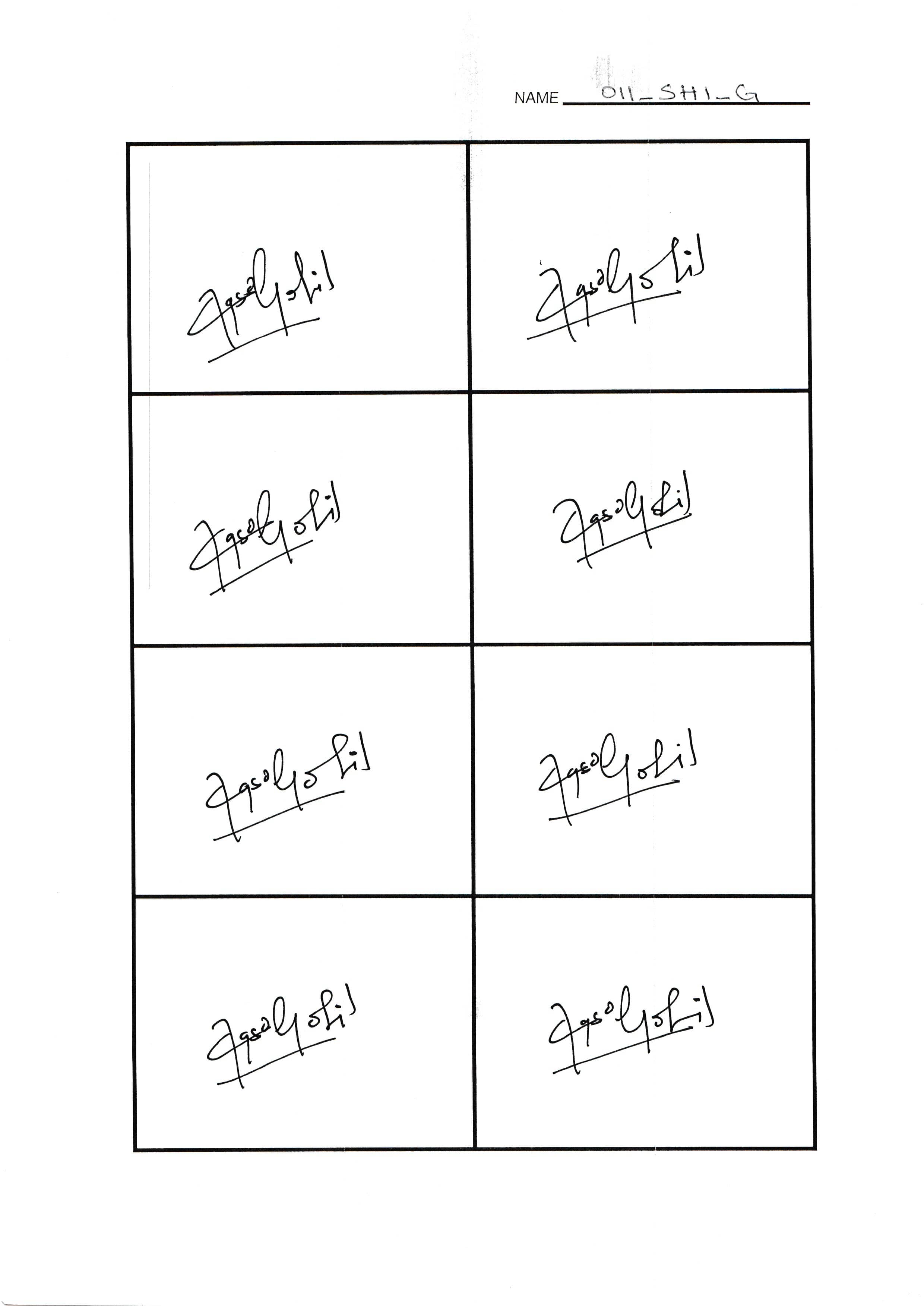}
     \end{subfigure}
     \begin{subfigure}[b]{0.325\textwidth}
         \centering
         \includegraphics[width=\textwidth]{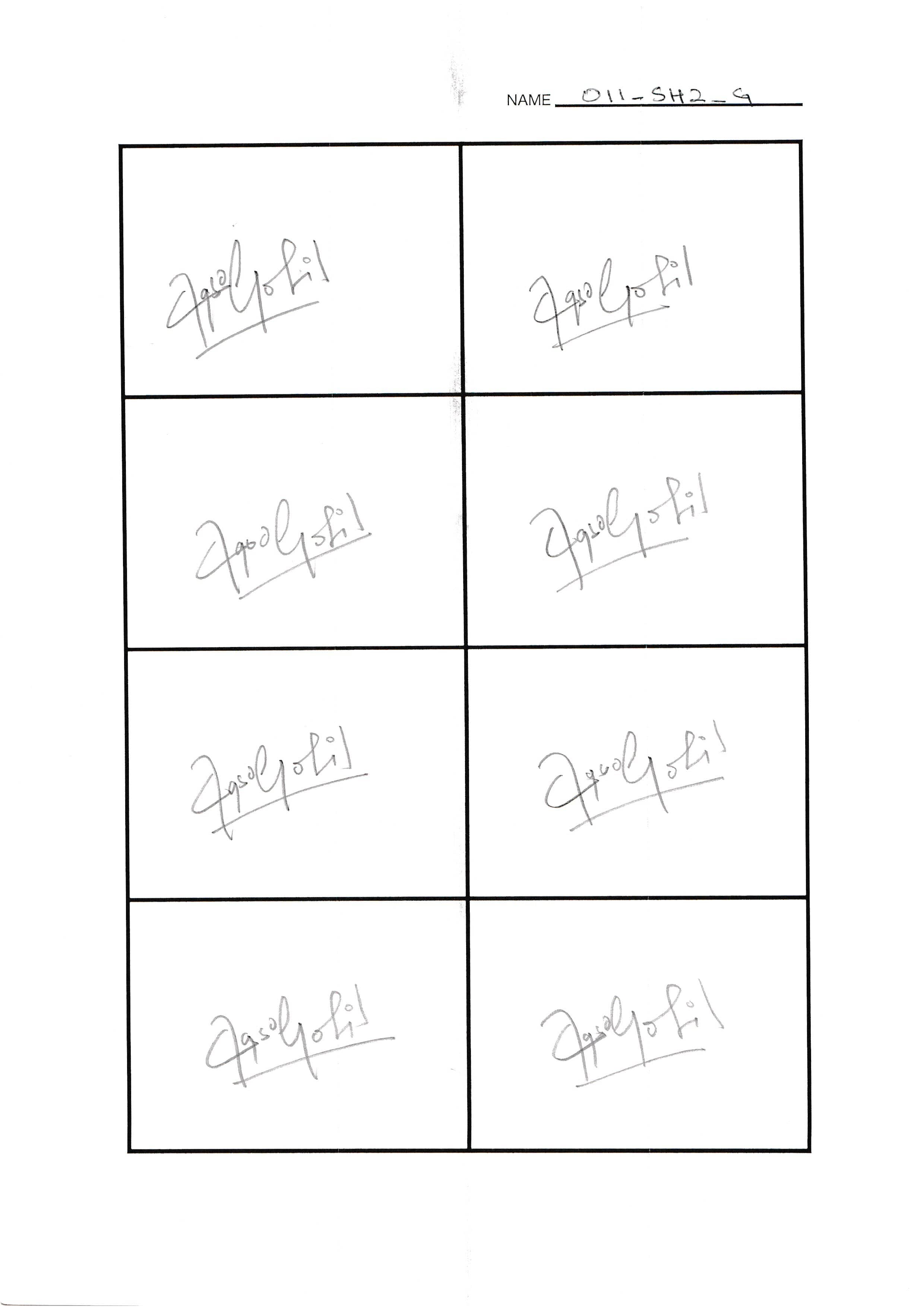}
     \end{subfigure}
     \begin{subfigure}[b]{0.325\textwidth}
         \centering
         \includegraphics[width=\textwidth]{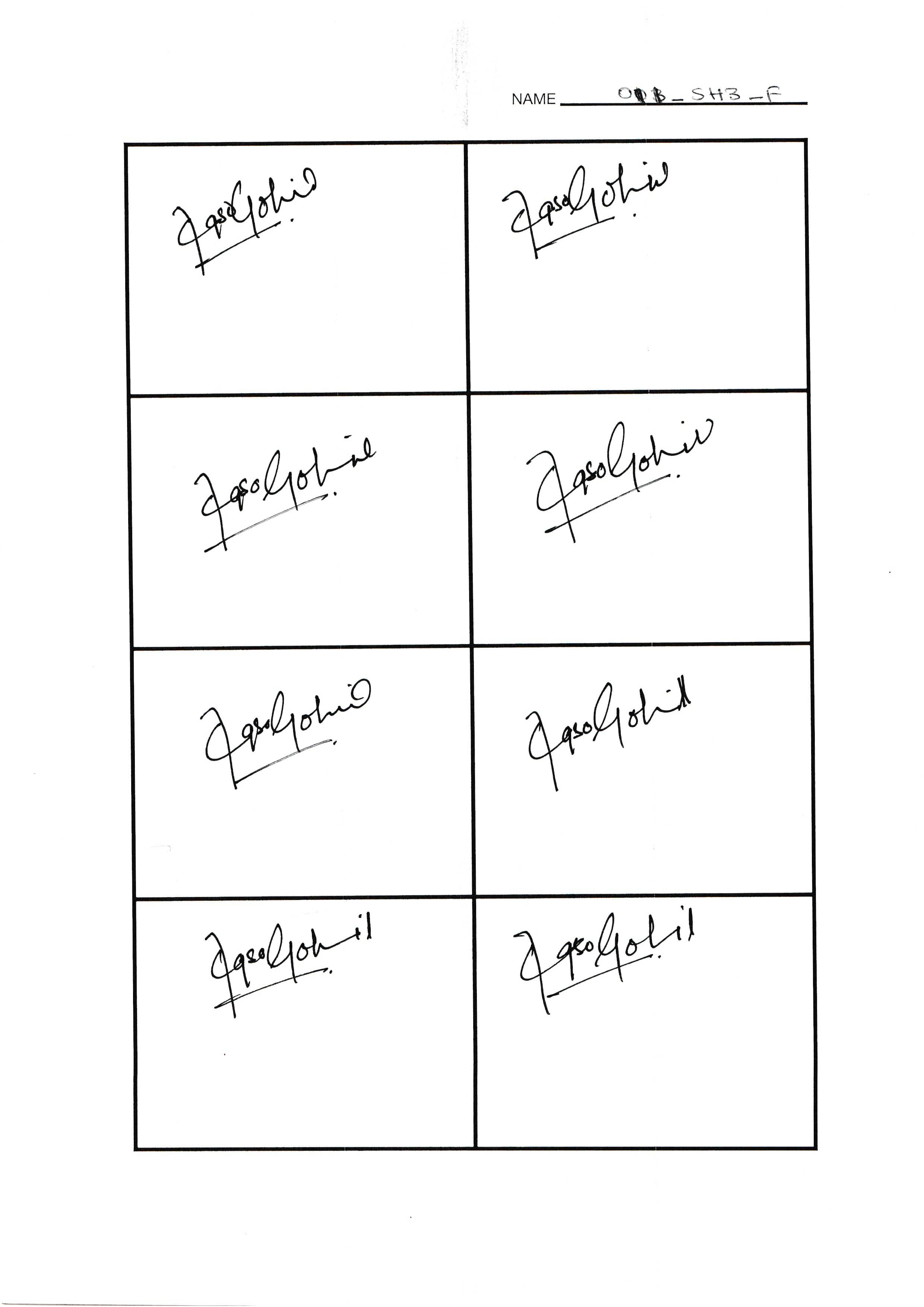}
     \end{subfigure}
    \caption{\textbf{Comparative Analysis of Signature Samples} - The two left columns display authentic signatures from an individual, showing the natural variations in their handwriting. The right column presents a forged signature sample, illustrating the outcome of a highly skilled forgery attempt with minimal visual differences from the genuine signatures. For each individual, eight forgeries were created: four executed with a ballpoint pen and four with a pencil to capture the diverse techniques used in forgery attempts.}
    \label{fig:sample_sheets}
\end{figure}

After collecting genuine signatures, we moved on to creating forgeries. We divided the signatures equally among three different people. These individuals were allowed to examine all authentic signatures of a person and practice them as long as they desired before creating the forgeries. As a result, we obtained highly skilled forgeries with very little evident visual differences from the genuine signatures. For each person, eight forgeries were created: four with a ballpoint pen and four with a pencil. In Fig.~\ref{fig:sample_sheets}, we show the raw signature samples collected for one individual.

% Details of genuine and forged samples collected for each person are provided in Table~\ref{tab:signature_details}, including a mention of the missing samples where forgeries were not available. 

% \begin{table}[ht]
%     \centering
%     \small
%     \begin{tabular}{ccc}
%         \toprule
%         & \multicolumn{2}{c}{\textbf{Num. Signatures}} \\
%         \cmidrule{2-3}
%         \textbf{Person} & \textbf{Genuine} & \textbf{Forgeries} \\
%         \cmidrule(l){1-1} \cmidrule(l){2-2} \cmidrule(l){3-3}
%         1 & 16 & 8 \\
%         2 & -  & - \\
%         3 & -  & - \\
%         4 & -  & - \\
%         5 & -  & - \\
%         6 & -  & - \\
%         7 & -  & - \\
%         8 & -  & - \\
%         9 & -  & - \\
%         10 & -  & - \\
%         11 & -  & - \\
%         12 & -  & - \\
%         13 & -  & - \\
%         14 & -  & - \\
%         15 & -  & - \\
%         16 & -  & - \\
%         17 & -  & - \\
%         \bottomrule
%     \end{tabular}
%     \caption{Caption}
%     \label{tab:signature_details}
% \end{table}

After obtaining the raw signatures, we manually drew bounding boxes around them using an annotation tool. We then used these annotations to crop the signatures from the collection sheets.

\subsection{Bank Check Creation}
\begin{figure}
\centering
\includegraphics[width=0.85\textwidth]{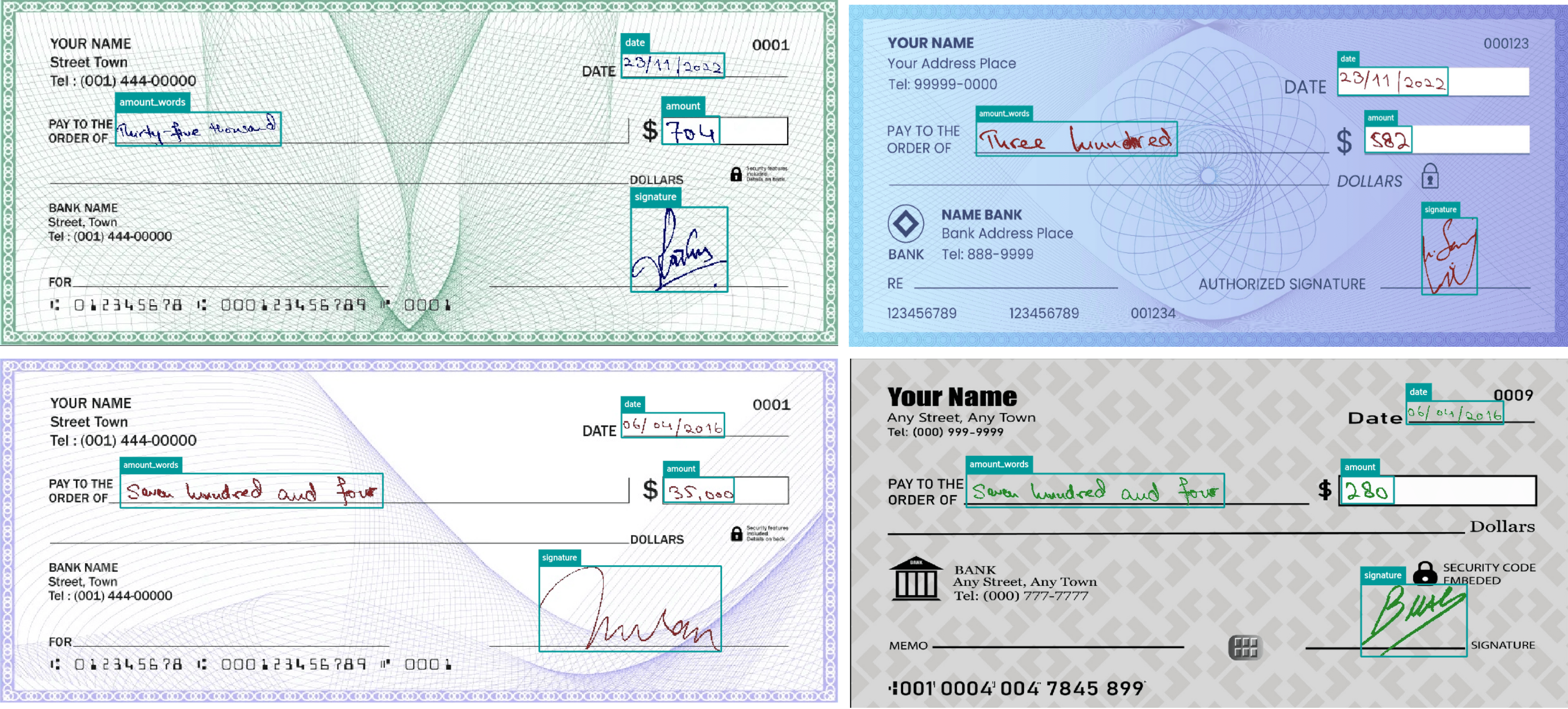}
\caption{Our Bank Check Data Samples offer a diverse collection of check designs and ink colors, mimicking the variety banks handle daily. With everything from basic blue and green to intricate patterns, it challenges signature verification by altering signature visibility. The range of ink colors tests detection capabilities across different contrasts. Essential for creating algorithms that accurately detect forgeries, this dataset is key to enhancing transaction security.}\label{fig:data_gt}
\end{figure}
In the second step, we obtain ten high-quality and realistic images of bank checks. These checks had varying degrees of layout and background complexity. We identify specific regions where different check elements are typically present for each check, such as the name, amount, date, and signature. After that, we use an automated algorithm to fill these regions with the collected signatures and some handwritten fake names, dates, and amounts. We will describe this algorithm in detail below.
We use annotated bounding boxes to crop the signatures from collection sheets and then apply a threshold to obtain a binary segmentation mask. We copy the signature pixels from the collection sheet and then augment them with a random pen color ink selected from a pool of common ink colors. These colors include black, dark gray, dark blue, red, and green, with probabilities of 0.2, 0.2, 0.3, 0.2, and 0.1, respectively. Five such augmentations are created. Next, we randomly select five checks from the checks database. Colored signature pixels are blended with the selected check background and positioned inside the signature area identified on the check. We also randomly scale and translate the signatures within the signature area, such that augmented signatures vary in location, size, and color. After filling in the signature field, the remaining check fields are filled similarly using fake data. It aims to give the bank check a realistic and ``filled'' appearance. Some examples of our bank checks are shown in~\cref{fig:data_gt}. 

\begin{table}[ht]
\centering
\caption{Detailed data splits for genuine and forged bank checks.}
\begin{tabular}{lccc}
\toprule
& \textbf{Train} & \textbf{Validation} & \textbf{Total} \\
\midrule
Number of genuine bank checks & 2352 & 1008 & 3360 \\
Number of forged bank checks & 700 & 300 & 1000 \\
\midrule
Total per split & 3052 & 1308 & 4360 \\
\bottomrule
\end{tabular}
\label{tab:data_split}
\end{table}
We provide labeled bounding boxes for the signature, legal amount, courtesy amount, date, and payee. The annotations include the ink color, a person ID to identify the signature author, and a boolean value indicating whether the signature is genuine or forged.
\subsection{Dataset Description}
We provide our dataset annotations in COCO format and images generated via our semi-automated check creation pipeline. Our dataset contains 4360 samples, with 3360 for training and the other 1000 for validation. We annotate six unique classes on the check, including the courtesy amount, legal amount, date, payee, and signature. The signature class is subdivided into genuine and forged signatures. Table~\ref{tab:data_split} shows the detailed split information including the number of checks in each split containing either a genuine or forged signature.
\section{Methodology}
\label{sec:method}

\subsection{Pre-processing}
\label{sec:preprocessig}
\begin{figure}
\centering
\includegraphics[width=0.95\textwidth]{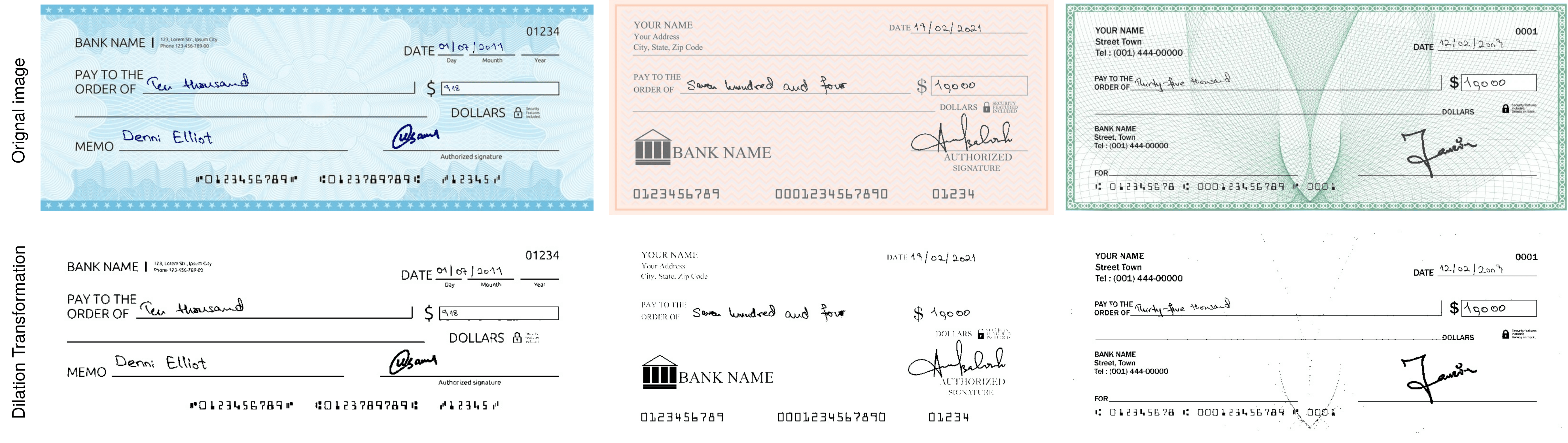}
\caption{Illustration of bank checks before and after a dilation transformation}\label{fig:bankcheque-dilation}
\end{figure}

\textbf{Dilation Transformation}
To enhance the accuracy of signature detection on bank checks, we apply a dilation transformation to the scanned images. This operation expands the pixels in the signature area, making faint or thin lines more prominent. This enhancement is particularly beneficial for lightly written or finely stroked signatures, improving their visibility and making them easier to detect. This pre-processing step is crucial for preparing the bank check data, ensuring that the signature detection system can accurately identify and analyze the signatures, as illustrated in Fig.~\ref{fig:bankcheque-dilation}.
\subsection{Network Architecture}
Our approach features an end-to-end architecture designed for detecting various fields on bank check images, as shown in Fig.\ref{fig:dia}. The network comprises two main modules: the training and guiding modules. Both modules utilize an ImageNet~\cite{ImageNet8} pre-trained ResNet-50 backbone\cite{resnet45} integrated with a transformer encoder-decoder network~\cite{shehzadi2023object5}. The detailed descriptions of these modules are as follows:

\begin{figure}
\centering
\includegraphics[width=0.9\textwidth]{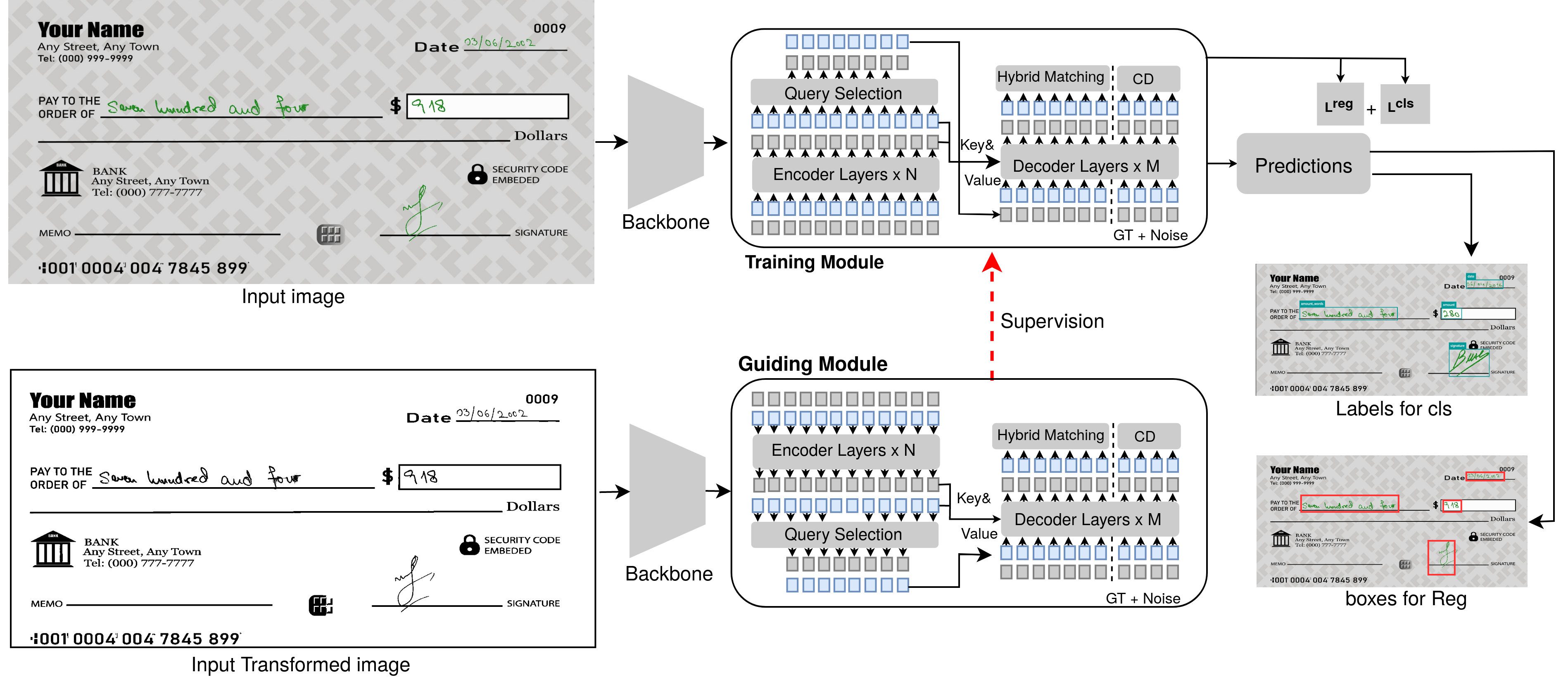}
\caption{
Overview of our proposed approach for bank checks. The process begins with the input of two cashier's checks, each with a signature. The top check is fed into a Training Module, which consists of a DINO network~\cite{dino23} with query selection, encoder, and decoder layers, and a hybrid matching component, where the model learns to predict the authenticity of the signature through supervised learning. After dilation transformation, the bottom check is processed through a guiding module, which parallels the Training Module's architecture but focuses on guiding the training process toward more stable and generalizable feature extraction.}\label{fig:dia}
\end{figure}
\begin{itemize}
    \item \textbf{Training Module:} This module uses the ResNet-50 backbone pre-trained on ImageNet to extract multi-scale features from the input images. The encoder enhances these features using positional embeddings derived from convolution layers with a 3x3 kernel size. A key feature of the DINO network is its mixed query selection strategy, which initializes positional queries and anchors while keeping content queries adaptable and advantageous during domain shifts. This strategy also supports Contrastive Denoising Training (CDN). In the decoder, deformable attention is employed to integrate the encoder's output with sequential query updates. CDN helps identify and rectify misinterpreted areas by passing gradients between adjacent layers early in the process. The final output for bank check images is obtained by calculating the dot product between the final query embedding and the pixel embedding map. The training module's performance is evaluated during inference using unseen data, excluding the guiding module.
    \item \textbf{Guiding Module:} This module is designed to enhance the quality of input bank checks by applying a dilation transformation, improving the training module's performance. It also employs the DINO network as a baseline. During the training phase, the guiding module actively updates and adjusts the training module, ensuring optimal adaptation and learning efficiency. This dynamic interaction between the two modules significantly enhances the system's overall effectiveness, especially in accurately processing and interpreting bank check images. For more details, please refer to the DINO network~\cite{dino23}.
\end{itemize}
Moreover, we propose a novel formulation for \textbf{writer-independent signature verification} using the detection network. This approach involves treating genuine and forged signatures as two distinct classes, training the network to classify detected signatures into one of these categories. Writer-independent verification means the system can authenticate signatures regardless of the specific writer, focusing on identifying common features of forgery across different signatures. This formulation allows the network to both locate signatures on bank checks and verify their authenticity by distinguishing between genuine and forged signatures, as demonstrated in our experiments.

\section{Experiments}
\label{sec:Experiments}

This section provides a detailed evaluation of our approach for detecting bank check fields, including verification of detected signatures. We also present some ablations to study the impact of different network components.

\subsection{Experimental Setup}

As a baseline for comparison, we utilize the DINO network~\cite{dino23} with a ResNet-50 backbone. We resize an input image, ensuring its shorter side falls between 480 and 800 pixels and its longer side is no more than 1333 pixels. Our training process involves training the networks for five epochs on Nvidia A100 GPUs with a batch size of 16. We use the AdamW optimizer with a $1e^{-4}$ weight decay. During training, we used 90 object queries in the training and guiding network decoder. Our model comprises a 6-layer transformer encoder and decoder with 256 hidden dimensions.

\begin{table}
    \centering
    \caption{Distribution of small, medium, and large annotations across the different dataset classes.}
    \begin{tabular}{ccccccc}
    \toprule
                   & \multicolumn{3}{c}{\textbf{Training}} & \multicolumn{3}{c}{\textbf{Validation}} \\
                   \cmidrule(l){2-4} \cmidrule(l){5-7}
    \textbf{Class} & \textbf{Small} & \textbf{Medium} & \textbf{Large} & \textbf{Small} & \textbf{Medium} & \textbf{Large} \\
    \midrule
    Amount (Courtesy) & 440 & 1881 &  731 & 169 & 813 & 326 \\
    Amount (Legal)    &   0 & 1142 & 1910 &   0 & 481 & 827 \\
    Date              & 163 & 1837 &  754 &  57 & 777 & 334 \\
    Payee             &   0 &  994 &  717 &   0 & 387 & 300 \\
    Signature (F)     &   2 &  304 &  394 &   0 & 123 & 177 \\
    Signature (G)     &   4 &  915 & 1433 &   1 & 358 & 649 \\
    \bottomrule
    \end{tabular}
    \label{tab:dataset_statistics}
\end{table}

\subsection{Evaluation Metrics}

We report the mean Average Precision (\(mAP\)) at the  Intersection over Union (IoU) threshold range of [0.5, 0.95], which gives the network overall accuracy in identifying various check components. Additionally, we analyze Average Precision (\(AP_{S}\), \(AP_{M}\), \(AP_{L}\)) metrics, with IoU threshold of 0.5, to evaluate the system's precision in detecting objects of small, medium, and large sizes, ensuring accurate identification of all critical elements on the check. Table~\ref{tab:dataset_statistics} provides a detailed breakdown of annotations by size across different classes. Alongside, Average Recall (\(AR_{S}\), \(AR_{M}\), \(AR_{L}\)) metrics, with IoU threshold of 0.5, are utilized to measure the system's capability to consistently detect relevant objects across different sizes, highlighting the model's efficiency in capturing a wide range of check features. This comprehensive set of metrics ensures that our bank check processing systems are both secure by effectively verifying signatures and efficient by precisely detecting and classifying check details.

\subsection{Results and Discussions}

This section presents the results of our main experiments, including baseline comparisons. The overall performance of our model for detecting bank check fields is evaluated in Table~\ref{tab:baseline}. A more detailed, class-wise comparison is provided in Table~\ref{tab:class_wise}, including our approach to distinguishing between genuine and forged signatures in a writer-independent verification setup.

\begin{table*}[ht]
\centering
\caption{Performance comparison with the baseline for detecting objects on bank checks}
\renewcommand{\arraystretch}{1}
\begin{tabular*}{.95\textwidth}
{@{\extracolsep{\fill}}lccccccc@{\extracolsep{\fill}}}
\toprule
\textbf{Architecture} & 
\textbf{mAP} & 
\textbf{AP\textsubscript{S}} &
\textbf{AP\textsubscript{M}} &
\textbf{AP\textsubscript{L}} &
\textbf{AR\textsubscript{S}} &
\textbf{AR\textsubscript{M}} &
\textbf{AR\textsubscript{L}} \\
\toprule
\multirow{1}{*}{DINO} & 94.5 & 65.2 & 94.6 & 95.1 & 83.3 & 97.0 & 97.8 \\
\multirow{1}{*}{Our}   & 99.7 & 96.7 & 99.8 & 99.7 & 96.7 & 99.9 & 99.8  \\ 
\bottomrule
\end{tabular*}
\label{tab:baseline}
\end{table*}

\subsubsection*{Baseline Comparisons:}
Our approach is also compared against the baseline DINO model, focusing specifically on bank check objects as shown in Table~\ref{tab:baseline}. The baseline DINO network achieved a mean Average Precision (mAP) score of 94.5. In contrast, our enhanced method significantly improved, reaching a mAP score 99.7. This improvement is especially notable in detecting small objects common on bank checks. Here, our method achieved an Average Precision (AP) of 96.7 and an Average Recall (AR) of 96.7, compared to the baseline's 65.2 AP and 83.3 AR. These results underscore the superior capability of our approach to accurately detect objects on bank checks, which is crucial for effective check analysis.

\begin{table*}[ht]
\centering
\caption{Class-wise performance evaluation for detecting and verifying objects on bank checks, including genuine and forged signatures, dates, monetary amounts, and payees.}
\renewcommand{\arraystretch}{1.2} % Adjusted for better spacing
\begin{tabular*}{0.98\textwidth}{@{\extracolsep{\fill}}lccccccc@{\extracolsep{\fill}}}
\toprule
\multirow{2}{*}{\textbf{Method}} & \multicolumn{2}{c}{\textbf{Signature}} & \multirow{2}{*}{\textbf{Date}} & \multicolumn{2}{c}{\textbf{Amount}} & \multirow{2}{*}{\textbf{Payee}} & \multirow{2}{*}{\textbf{Overall}} \\
\cmidrule{2-3} \cmidrule{5-6}
& \textbf{Genuine} &  \textbf{Forged} & & \textbf{Courtesy} & \textbf{Legal} & & \\
\midrule
\textbf{DINO} & 93.1 & 89.3 & 93.7 & 96.6 & 97.4 & 97.2 & 94.5\\
\textbf{Our}  & 99.2 & 99.4 & 100  & 100  & 100 &  99.7 & 99.7 \\
\bottomrule
\end{tabular*}
\label{tab:class_wise}
\end{table*}

\subsubsection*{Performance in Signature Verification:}
We conducted an in-depth analysis of our system's performance in verifying signatures on bank checks. Our method demonstrates significant improvements over the baseline DINO model, achieving an Average Precision (AP) of 99.2 for genuine signatures and 99.4 for forged signatures. In contrast, the DINO baseline achieved 93.1 AP for genuine and 89.3 AP for forged signatures. This marked improvement underscores the robustness of our detection-based verification approach, which effectively reduces false positives and negatives, ensuring higher reliability in fraud detection. These results validate our claims in the abstract and introduction, showcasing the system's capability to handle real-world complexities in bank check signatures.

\subsubsection*{Detection of Other Bank Check Elements:}
Our method also excels in detecting other critical elements on bank checks, including dates, monetary amounts (courtesy and legal), and payees. The AP for these fields reached 100 with our approach, compared to the DINO baseline's 93.7, 96.6, 97.4, and 97.2, respectively. This exceptional performance indicates our network's ability to accurately identify and localize various components on the checks, further enhancing the system's overall reliability. The results, as presented in Table~\ref{tab:class_wise}, highlight the comprehensive improvement achieved by our framework across all evaluated components of bank checks.

\subsubsection*{Implications for Real-World Applications:}
Our findings demonstrate the practical utility of our approach in real-world banking operations. The substantial improvements in both signature verification and the detection of other check elements suggest that our method can significantly enhance the security and efficiency of financial document processing. Our work sets the stage for future advancements in secure and efficient signature verification technologies by providing a robust dataset and a powerful detection network.

\subsection{Ablation Studies}

\subsubsection*{Impact of the Guiding Module:}
The guiding module's contribution to the network's performance is illustrated in Table~\ref{tab:train_guide}. Incorporating the guiding module results in an increase in mAP from 94.5 to 97.1. The integration of a guiding module with a dilation transformation significantly boosts feature extraction, enhancing the visibility of thin strokes and faint signatures. This pre-processing step is crucial for improving detection accuracy. 
\begin{table*}[ht]
\centering
\caption{Performance comparison with and without the guiding module and pre-processing step. Here, pre-processing is applied to the guiding module.}
\renewcommand{\arraystretch}{1}
\begin{tabular*}{.95\textwidth}
{@{\extracolsep{\fill}}cccccccccc@{\extracolsep{\fill}}}
\toprule
\textbf{Training} & 
\textbf{Guiding} & 
\textbf{Pre-processing} &
\textbf{mAP} & 
\textbf{AP\textsubscript{S}} &
\textbf{AP\textsubscript{M}} &
\textbf{AP\textsubscript{L}} &
\textbf{AR\textsubscript{S}} &
\textbf{AR\textsubscript{M}} &
\textbf{AR\textsubscript{L}} \\
\toprule
\multirow{1}{*}{\checkmark } &  \color{black}\xmark  & - & 94.5 & 65.2 & 94.6 & 95.1 & 83.3 & 97.0 & 97.8 \\
\multirow{1}{*}{\checkmark } &  \checkmark  & - & 97.1 & 90.8 & 97.5 & 97.2 & 91.9 & 98.4 & 98.3 \\
             \checkmark   & \checkmark & Dilation  & 99.7 & 96.7 & 99.8 & 99.7 & 96.7 & 99.9 & 99.8 \\
\bottomrule
\end{tabular*}
\label{tab:train_guide}
\vspace{-15pt}
\end{table*}

\subsubsection*{Impact of Dilation Transformation:}
The positive effects of applying dilation transformation in the guiding module are evident. This preprocessing step improves the network's overall performance, as shown in the last two rows of Table~\ref{tab:train_guide}. Dilation enhances the visibility of critical foreground information, such as text and signatures, by minimizing the interference of background patterns. This improvement is crucial for the training module to recognize important details on bank checks effectively. Moreover, dilation aids in verifying signatures by emphasizing key features, thus facilitating more reliable detection and verification of bank check elements.

\section{Conclusion}
\label{sec:conclusion}

In this paper, we addressed the critical task of signature verification on bank checks, which is essential for preventing fraud and ensuring transaction authenticity. We introduced the Synthetic Signature Bankcheck Images (SSBI) dataset, a novel collection of signatures in complex scenarios, including real and forged signatures embedded within typical check elements. This dataset provides a realistic and challenging environment for advancing signature detection and verification methods.

Furthermore, we presented an end-to-end trainable framework based on the DINO architecture, augmented with a dilation module to enhance the detection and verification of signatures on bank checks. Our method demonstrated significant improvements, achieving a notable increase in performance metrics over the baseline DINO network. Specifically, our approach achieved an mAP of 99.7, with substantial gains in detecting small objects, underscoring the effectiveness of our guiding module and dilation preprocessing.

Our results highlight the potential of our framework to improve the accuracy and reliability of signature verification systems, which is crucial for enhancing security and operational efficiency in financial document processing. The SSBI dataset, along with our proposed methodology, lays the groundwork for future research in developing robust and advanced techniques to combat signature forgery and improve fraud detection in banking and other domains where signature verification is vital.

By providing a comprehensive dataset and a powerful detection-based verification approach, this work contributes significantly to the field of automated signature verification, offering a practical solution for real-world applications and setting the stage for further advancements in secure document authentication.

\section*{Acknowledgements}
The work leading to this publication has been partially funded by the EU Horizon Europe Project AIRISE (https://airise.eu/) under grant agreement 101092312.

% ---- Bibliography ----
%
% BibTeX users should specify bibliography style 'splncs04'.
% References will then be sorted and formatted in the correct style.
%
% \bibliographystyle{splncs04}
% \bibliography{mybibliography}
%
\bibliographystyle{IEEEtran}
\bibliography{main}% common bib file

% Generated by IEEEtran.bst, version: 1.14 (2015/08/26)
\begin{thebibliography}{10}
\providecommand{\url}[1]{#1}
\csname url@samestyle\endcsname
\providecommand{\newblock}{\relax}
\providecommand{\bibinfo}[2]{#2}
\providecommand{\BIBentrySTDinterwordspacing}{\spaceskip=0pt\relax}
\providecommand{\BIBentryALTinterwordstretchfactor}{4}
\providecommand{\BIBentryALTinterwordspacing}{\spaceskip=\fontdimen2\font plus
\BIBentryALTinterwordstretchfactor\fontdimen3\font minus \fontdimen4\font\relax}
\providecommand{\BIBforeignlanguage}[2]{{%
\expandafter\ifx\csname l@#1\endcsname\relax
\typeout{** WARNING: IEEEtran.bst: No hyphenation pattern has been}%
\typeout{** loaded for the language `#1'. Using the pattern for}%
\typeout{** the default language instead.}%
\else
\language=\csname l@#1\endcsname
\fi
#2}}
\providecommand{\BIBdecl}{\relax}
\BIBdecl

\bibitem{khan2018signature}
\BIBentryALTinterwordspacing
M.~S.~U. Khan, M.~M. Tariq, and B.~Ahmad, ``Signature verification,'' 2018. [Online]. Available: \url{https://www.researchgate.net/publication/339299291_Signature_Verification}
\BIBentrySTDinterwordspacing

\bibitem{dargan2020survey}
\BIBentryALTinterwordspacing
S.~Dargan and M.~Kumar, ``A comprehensive survey on the biometric recognition systems based on physiological and behavioral modalities,'' \emph{Expert Systems with Applications}, vol. 143, p. 113114, 2020. [Online]. Available: \url{https://www.sciencedirect.com/science/article/pii/S0957417419308310}
\BIBentrySTDinterwordspacing

\bibitem{liang2020behavioral}
Y.~Liang, S.~Samtani, B.~Guo, and Z.~Yu, ``Behavioral biometrics for continuous authentication in the internet-of-things era: An artificial intelligence perspective,'' \emph{IEEE Internet of Things Journal}, vol.~7, no.~9, pp. 9128--9143, 2020.

\bibitem{sarkar2020review}
A.~Sarkar and B.~K. Singh, ``A review on performance, security and various biometric template protection schemes for biometric authentication systems,'' \emph{Multimedia Tools and Applications}, vol.~79, pp. 27\,721--27\,776, 2020.

\bibitem{Kao_app101}
\BIBentryALTinterwordspacing
H.-H. Kao and C.-Y. Wen, ``An offline signature verification and forgery detection method based on a single known sample and an explainable deep learning approach,'' \emph{Applied Sciences}, vol.~10, no.~11, 2020. [Online]. Available: \url{https://www.mdpi.com/2076-3417/10/11/3716}
\BIBentrySTDinterwordspacing

\bibitem{Vorugunti_FuseNetOS}
\BIBentryALTinterwordspacing
C.~S. Vorugunti, V.~Pulabaigari, R.~K. S.~S. Gorthi, and P.~Mukherjee, ``Osvfusenet: Online signature verification by feature fusion and depth-wise separable convolution based deep learning,'' \emph{Neurocomputing}, vol. 409, pp. 157--172, 2020. [Online]. Available: \url{https://api.semanticscholar.org/CorpusID:221381079}
\BIBentrySTDinterwordspacing

\bibitem{pal_sig34}
S.~Pal, A.~Alaei, U.~Pal, and M.~Blumenstein, ``Performance of an off-line signature verification method based on texture features on a large indic-script signature dataset,'' in \emph{2016 12th IAPR Workshop on Document Analysis Systems (DAS)}, 2016, pp. 72--77.

\bibitem{OKAWA_2018480}
\BIBentryALTinterwordspacing
M.~Okawa, ``Synergy of foreground–background images for feature extraction: Offline signature verification using fisher vector with fused kaze features,'' \emph{Pattern Recognition}, vol.~79, pp. 480--489, 2018. [Online]. Available: \url{https://www.sciencedirect.com/science/article/pii/S0031320318300803}
\BIBentrySTDinterwordspacing

\bibitem{FIERREZ20}
\BIBentryALTinterwordspacing
J.~Fierrez, J.~Ortega-Garcia, D.~Ramos, and J.~Gonzalez-Rodriguez, ``Hmm-based on-line signature verification: Feature extraction and signature modeling,'' \emph{Pattern Recognition Letters}, vol.~28, no.~16, pp. 2325--2334, 2007. [Online]. Available: \url{https://www.sciencedirect.com/science/article/pii/S0167865507002395}
\BIBentrySTDinterwordspacing

\bibitem{patel_2017}
\BIBentryALTinterwordspacing
S.~D. Das, H.~Ladia, V.~Kumar, and S.~Mishra, ``Writer independent offline signature recognition using ensemble learning,'' \emph{CoRR}, vol. abs/1901.06494, 2019. [Online]. Available: \url{http://arxiv.org/abs/1901.06494}
\BIBentrySTDinterwordspacing

\bibitem{Narwade_2018OfflineSV}
\BIBentryALTinterwordspacing
P.~N. Narwade, R.~R. Sawant, and S.~V. Bonde, ``Offline signature verification using shape correspondence,'' \emph{Int. J. Biom.}, vol.~10, pp. 272--289, 2018. [Online]. Available: \url{https://api.semanticscholar.org/CorpusID:67868525}
\BIBentrySTDinterwordspacing

\bibitem{dino23}
\BIBentryALTinterwordspacing
H.~Zhang, F.~Li, S.~Liu, L.~Zhang, H.~Su, J.~Zhu, L.~M. Ni, and H.-Y. Shum, ``Dino: Detr with improved denoising anchor boxes for end-to-end object detection,'' 2022. [Online]. Available: \url{https://arxiv.org/abs/2203.03605}
\BIBentrySTDinterwordspacing

\bibitem{Fierrez_MCYT-75}
\BIBentryALTinterwordspacing
J.~Fierrez, L.~Nanni, J.~Lopez-Pe{\~n}alba, J.~Ortega-Garcia, and D.~Maltoni, ``An on-line signature verification system based on fusion of local and global information,'' in \emph{International Conference on Audio- and Video-Based Biometric Person Authentication}, 2005. [Online]. Available: \url{https://api.semanticscholar.org/CorpusID:2607577}
\BIBentrySTDinterwordspacing

\bibitem{SHARIF_202050}
\BIBentryALTinterwordspacing
M.~Sharif, M.~A. Khan, M.~Faisal, M.~Yasmin, and S.~L. Fernandes, ``A framework for offline signature verification system: Best features selection approach,'' \emph{Pattern Recognition Letters}, vol. 139, pp. 50--59, 2020. [Online]. Available: \url{https://www.sciencedirect.com/science/article/pii/S016786551830028X}
\BIBentrySTDinterwordspacing

\bibitem{GPDS_data56}
M.~A. Ferrer, M.~Diaz-Cabrera, and A.~Morales, ``Synthetic off-line signature image generation,'' in \emph{2013 International Conference on Biometrics (ICB)}, 2013, pp. 1--7.

\bibitem{shehzadi_IEEE_I9}
T.~Shehzadi, A.~Majid, M.~Hameed, A.~Farooq, and A.~Yousaf, ``Intelligent predictor using cancer-related biologically information extraction from cancer transcriptomes,'' in \emph{2020 International Symposium on Recent Advances in Electrical Engineering \& Computer Sciences (RAEE \& CS)}, vol.~5, 2020, pp. 1--5.

\bibitem{Protein10}
A.~Yousaf, T.~Shehzadi, A.~Farooq, and K.~Ilyas, ``Protein active site prediction for early drug discovery and designing,'' \emph{International Review of Applied Sciences and Engineering}, vol.~13, no.~1, pp. 98--105, 2021.

\bibitem{wajahatCC8}
W.~Saeed, M.~S. Saleh, M.~N. Gull, H.~Raza, R.~Saeed, and T.~Shehzadi, ``Geometric features and traffic dynamic analysis on 4-leg intersections,'' \emph{International Review of Applied Sciences and Engineering}, 2023.

\bibitem{semi-detr_table1}
T.~Shehzadi, K.~Azeem~Hashmi, D.~Stricker, M.~Liwicki, and M.~Zeshan~Afzal, ``Towards end-to-end semi-supervised table detection with deformable transformer,'' in \emph{Document Analysis and Recognition - ICDAR 2023}, G.~A. Fink, R.~Jain, K.~Kise, and R.~Zanibbi, Eds.\hskip 1em plus 0.5em minus 0.4em\relax Cham: Springer Nature Switzerland, 2023, pp. 51--76.

\bibitem{sunsupdla12}
T.~U. Sheikh, T.~Shehzadi, K.~A. Hashmi, D.~Stricker, and M.~Z. Afzal, ``Unsupdla: Towards unsupervised document layout analysis,'' 2024.

\bibitem{shehzadi2024hybrid6}
T.~Shehzadi, D.~Stricker, and M.~Z. Afzal, ``A hybrid approach for document layout analysis in document images,'' 2024.

\bibitem{shehzadi2024endtoend7}
T.~Shehzadi, S.~Sarode, D.~Stricker, and M.~Z. Afzal, ``Towards end-to-end semi-supervised table detection with semantic aligned matching transformer,'' 2024.

\bibitem{ehsan_semi8}
I.~Ehsan, T.~Shehzadi, D.~Stricker, and M.~Z. Afzal, ``End-to-end semi-supervised approach with modulated object queries for table detection in documents,'' \emph{arXiv preprint arXiv:2405.04971}, 2024.

\bibitem{continuaLR45}
\BIBentryALTinterwordspacing
M.~Minouei, K.~A. Hashmi, M.~R. Soheili, M.~Z. Afzal, and D.~Stricker, ``Continual learning for table detection in document images,'' \emph{Applied Sciences}, vol.~12, no.~18, 2022. [Online]. Available: \url{https://www.mdpi.com/2076-3417/12/18/8969}
\BIBentrySTDinterwordspacing

\bibitem{Real_DICls4}
A.~Kölsch, M.~Z. Afzal, M.~Ebbecke, and M.~Liwicki, ``Real-time document image classification using deep cnn and extreme learning machines,'' in \emph{2017 14th IAPR International Conference on Document Analysis and Recognition (ICDAR)}, vol.~01, 2017, pp. 1318--1323.

\bibitem{cas10}
\BIBentryALTinterwordspacing
K.~A. Hashmi, A.~Pagani, M.~Liwicki, D.~Stricker, and M.~Z. Afzal, ``Cascade network with deformable composite backbone for formula detection in scanned document images,'' \emph{Applied Sciences}, vol.~11, no.~16, 2021. [Online]. Available: \url{https://www.mdpi.com/2076-3417/11/16/7610}
\BIBentrySTDinterwordspacing

\bibitem{sparse_semi_detr2}
T.~Shehzadi, K.~A. Hashmi, D.~Stricker, and M.~Z. Afzal, ``Sparse semi-detr: Sparse learnable queries for semi-supervised object detection,'' \emph{arXiv preprint arXiv:2404.01819}, 2024.

\bibitem{bridging_per3}
T.~Shehzadi, K.~A. Hashmi, D.~Stricker, M.~Liwicki, and M.~Z. Afzal, ``Bridging the performance gap between detr and r-cnn for graphical object detection in document images,'' \emph{arXiv preprint arXiv:2306.13526}, 2023.

\bibitem{semimask4}
T.~Shehzadi, K.~A. Hashmi, A.~Pagani, M.~Liwicki, D.~Stricker, and M.~Z. Afzal, ``Mask-aware semi-supervised object detection in floor plans,'' \emph{Applied Sciences}, vol.~12, no.~19, 2022.

\bibitem{Shariatmadari_2019}
\BIBentryALTinterwordspacing
S.~Shariatmadari, S.~Emadi, and Y.~Akbari, ``Patch-based offline signature verification using one-class hierarchical deep learning,'' \emph{International Journal on Document Analysis and Recognition (IJDAR)}, vol.~22, pp. 375--385, 2019. [Online]. Available: \url{https://api.semanticscholar.org/CorpusID:199443408}
\BIBentrySTDinterwordspacing

\bibitem{CEDAR}
H.~Srinivasan, S.~N. Srihari, and M.~J. Beal, ``Machine learning for signature verification,'' in \emph{Computer Vision, Graphics and Image Processing: 5th Indian Conference, ICVGIP 2006, Madurai, India, December 13-16, 2006. Proceedings}.\hskip 1em plus 0.5em minus 0.4em\relax Springer, 2006, pp. 761--775.

\bibitem{wei_2019_1}
P.~Wei, H.~Li, and P.~Hu, ``Inverse discriminative networks for handwritten signature verification,'' in \emph{2019 IEEE/CVF Conference on Computer Vision and Pattern Recognition (CVPR)}, 2019, pp. 5757--5765.

\bibitem{BHSig260}
S.~Pal, A.~Alaei, U.~Pal, and M.~Blumenstein, ``Performance of an off-line signature verification method based on texture features on a large indic-script signature dataset,'' in \emph{2016 12th IAPR workshop on document analysis systems (DAS)}.\hskip 1em plus 0.5em minus 0.4em\relax IEEE, 2016, pp. 72--77.

\bibitem{Jain_2020HandwrittenSV}
\BIBentryALTinterwordspacing
A.~Jain, S.~K. Singh, and K.~P. Singh, ``Handwritten signature verification using shallow convolutional neural network,'' \emph{Multimedia Tools and Applications}, vol.~79, pp. 19\,993--20\,018, 2020. [Online]. Available: \url{https://api.semanticscholar.org/CorpusID:214808456}
\BIBentrySTDinterwordspacing

\bibitem{CVBLSig}
------, ``Handwritten signature verification using shallow convolutional neural network,'' \emph{Multimedia Tools and Applications}, vol.~79, pp. 19\,993--20\,018, 2020.

\bibitem{pal_2016skri}
S.~Pal, A.~Alaei, U.~Pal, and M.~Blumenstein, ``Performance of an off-line signature verification method based on texture features on a large indic-script signature dataset,'' in \emph{2016 12th IAPR Workshop on Document Analysis Systems (DAS)}, 2016, pp. 72--77.

\bibitem{PODDAR_2020610}
\BIBentryALTinterwordspacing
J.~Poddar, V.~Parikh, and S.~K. Bharti, ``Offline signature recognition and forgery detection using deep learning,'' \emph{Procedia Computer Science}, vol. 170, pp. 610--617, 2020, the 11th International Conference on Ambient Systems, Networks and Technologies (ANT) / The 3rd International Conference on Emerging Data and Industry 4.0 (EDI40) / Affiliated Workshops. [Online]. Available: \url{https://www.sciencedirect.com/science/article/pii/S1877050920305731}
\BIBentrySTDinterwordspacing

\bibitem{BAY2008346}
\BIBentryALTinterwordspacing
H.~Bay, A.~Ess, T.~Tuytelaars, and L.~{Van Gool}, ``Speeded-up robust features (surf),'' \emph{Computer Vision and Image Understanding}, vol. 110, no.~3, pp. 346--359, 2008, similarity Matching in Computer Vision and Multimedia. [Online]. Available: \url{https://www.sciencedirect.com/science/article/pii/S1077314207001555}
\BIBentrySTDinterwordspacing

\bibitem{harris1988combined}
C.~Harris and M.~Stephens, ``A combined corner and edge detector,'' \emph{Procedings of the Alvey Vision Conference}, pp. 147--151, 1988.

\bibitem{Ghosh_2020ARN}
\BIBentryALTinterwordspacing
R.~Ghosh, ``A recurrent neural network based deep learning model for offline signature verification and recognition system,'' \emph{Expert Syst. Appl.}, vol. 168, p. 114249, 2020. [Online]. Available: \url{https://api.semanticscholar.org/CorpusID:228903333}
\BIBentrySTDinterwordspacing

\bibitem{GHOSH_b202113}
\BIBentryALTinterwordspacing
S.~Ghosh, S.~Ghosh, P.~Kumar, E.~Scheme, and P.~P. Roy, ``A novel spatio-temporal siamese network for 3d signature recognition,'' \emph{Pattern Recognition Letters}, vol. 144, pp. 13--20, 2021. [Online]. Available: \url{https://www.sciencedirect.com/science/article/pii/S0167865521000258}
\BIBentrySTDinterwordspacing

\bibitem{FCNRLAF7}
\BIBentryALTinterwordspacing
C.~A. M.~L. Junior, M.~H.~M. da~Silva, B.~L.~D. Bezerra, B.~J.~T. Fernandes, and D.~Impedovo, ``Fcn+rl: A fully convolutional network followed by refinement layers to offline handwritten signature segmentation,'' \emph{2020 International Joint Conference on Neural Networks (IJCNN)}, pp. 1--7, 2020. [Online]. Available: \url{https://api.semanticscholar.org/CorpusID:219124088}
\BIBentrySTDinterwordspacing

\bibitem{Liu2021_OfflineSV}
\BIBentryALTinterwordspacing
L.~Liu, L.~Huang, F.~Yin, and Y.~Chen, ``Offline signature verification using a region based deep metric learning network,'' \emph{Pattern Recognit.}, vol. 118, p. 108009, 2021. [Online]. Available: \url{https://api.semanticscholar.org/CorpusID:235677030}
\BIBentrySTDinterwordspacing

\bibitem{UTSIg45}
A.~Soleimani, K.~Fouladi, and B.~N. Araabi, ``Utsig: A persian offline signature dataset,'' \emph{IET Biometrics}, vol.~6, no.~1, pp. 1--8, 2017.

\bibitem{icdar11_sig}
A.~Shahab, F.~Shafait, and A.~Dengel, ``Icdar 2011 robust reading competition challenge 2: Reading text in scene images,'' in \emph{2011 International Conference on Document Analysis and Recognition}, 2011, pp. 1491--1496.

\bibitem{liwicki2011signature}
M.~Liwicki, M.~I. Malik, C.~E. Van Den~Heuvel, X.~Chen, C.~Berger, R.~Stoel, M.~Blumenstein, and B.~Found, ``Signature verification competition for online and offline skilled forgeries (sigcomp2011),'' in \emph{2011 International conference on document analysis and recognition}.\hskip 1em plus 0.5em minus 0.4em\relax IEEE, 2011, pp. 1480--1484.

\bibitem{khan2021novel}
M.~S.~U. Khan, ``A novel segmentation dataset for signatures on bank checks,'' 2021.

\bibitem{ImageNet8}
J.~Deng, W.~Dong, R.~Socher, L.-J. Li, K.~Li, and L.~Fei-Fei, ``Imagenet: A large-scale hierarchical image database,'' in \emph{2009 IEEE Conference on Computer Vision and Pattern Recognition}, 2009, pp. 248--255.

\bibitem{resnet45}
\BIBentryALTinterwordspacing
C.~Szegedy, S.~Ioffe, and V.~Vanhoucke, ``Inception-v4, inception-resnet and the impact of residual connections on learning,'' \emph{CoRR}, vol. abs/1602.07261, 2016. [Online]. Available: \url{http://arxiv.org/abs/1602.07261}
\BIBentrySTDinterwordspacing

\bibitem{shehzadi2023object5}
T.~Shehzadi, K.~A. Hashmi, D.~Stricker, and M.~Z. Afzal, ``Object detection with transformers: A review,'' 2023.

\end{thebibliography}

\end{document}